\documentclass{article}

\usepackage{PRIMEarxiv}

\usepackage[utf8]{inputenc} 
\usepackage[T1]{fontenc}    
\usepackage{hyperref}       
\usepackage{url}            
\usepackage{booktabs}       
\usepackage{amsfonts}       
\usepackage{nicefrac}       
\usepackage{microtype}      
\usepackage{lipsum}
\usepackage{fancyhdr}       
\usepackage{graphicx}       
\graphicspath{{media/}}     
\usepackage{amsmath} 
\usepackage{multirow} 
\usepackage{stfloats} 
\usepackage[numbers]{natbib}
\bibpunct[, ]{(}{)}{;}{a}{}{,}
\renewcommand\bibfont{\fontsize{10}{12}\selectfont}

\usepackage{tabularx}
\usepackage{bm}

\title{Sensor-Adaptive Flood Mapping with Pre-trained Multi-Modal Transformers across SAR and Multispectral Modalities
}

\author{
  Tomohiro Tanaka \\
  Graduate school of Science \& Engineering,  \\
  Saitama University, Japan \\
  \texttt{t.tanaka.579@ms.saitama-u.ac.jp} 
  \And
  Narumasa Tsutsumida \\
  Graduate school of Science \& Engineering,  \\
  Saitama University, Japan \\
  \texttt{rsnaru.jp@gmail.com} 
}

\begin{document}
\maketitle

\begin{abstract}
Floods are increasingly frequent natural disasters causing extensive human and economic damage, highlighting the critical need for rapid and accurate flood inundation mapping. While remote sensing technologies have advanced flood monitoring capabilities, operational challenges persist: single-sensor approaches face weather-dependent data availability and limited revisit periods, while multi-sensor fusion methods require substantial computational resources and large-scale labeled datasets. To address these limitations, this study introduces a novel sensor-flexible flood detection methodology by fine-tuning Presto, a lightweight ($\sim$0.4M parameters) multi-modal pre-trained transformer that processes both Synthetic Aperture Radar (SAR) and multispectral (MS) data at the pixel level. Our approach uniquely enables flood mapping using SAR-only, MS-only, or combined SAR+MS inputs through a single model architecture, addressing the critical operational need for rapid response with whatever sensor data becomes available first during disasters. We evaluated our method on the Sen1Floods11 dataset against the large-scale Prithvi-100M baseline ($\sim$100M parameters) across three realistic data availability scenarios. The proposed model achieved superior performance with an F1 score of 0.896 and mIoU of 0.886 in the optimal sensor-fusion scenario, outperforming the established baseline. Crucially, the model demonstrated robustness by maintaining effective performance in MS-only scenarios (F1: 0.893) and functional capabilities in challenging SAR-only conditions (F1: 0.718), confirming the advantage of multi-modal pre-training for operational flood mapping. Our parameter-efficient, sensor-flexible approach offers an accessible and robust solution for real-world disaster scenarios requiring immediate flood extent assessment regardless of sensor availability constraints.

\end{abstract}


\section*{1 Introduction}
\label{Introduction}

Floods constitute one of the most prevalent and catastrophic natural disasters worldwide, resulting in extensive human and economic impacts \citep{douris2021atlas}. Accurate and timely flood extent mapping is essential for effective disaster response, facilitating damage assessment, secondary hazard mitigation, and recovery planning. Nevertheless, achieving precise and rapid flood delineation remains challenging due to the dynamic nature of flood events, limited accessibility to affected regions, and critical time constraints inherent in emergency response scenarios.

Remote sensing technology offers rapid Earth observation capabilities, making it effective for monitoring flood disasters \citep{oddo2019value}.
Synthetic Aperture Radar (SAR) enables all-weather observations but suffers from speckle, complicating interpretation and reducing accuracy \citep{shen2019inundation}. Multispectral (MS) sensor offers intuitive interpretation but are highly vulnerable to cloud cover \citep{lin2016review}. 
During flood events, cloud coverage makes optical data acquisition particularly problematic, forcing emergency responders to rely on whichever sensor data becomes available first.
The Sentinel missions provide open-access, high-resolution data but face temporal limitations that reduce operational effectiveness. Sentinel-1 (S1) delivers SAR data with 10-m resolution and 6-12 day revisit periods, while Sentinel-2 (S2) provides multispectral (MS) optical data with 10-60 m resolution and 5-10 day revisit cycles. These revisit intervals create substantial operational gaps, as flood events typically persist for only 2.5-3.5 days \citep{Tarpanelli2022-ac}. Even under optimal conditions, S1 detects only 58\% of flood events while S2 captures merely 28\% due to cloud coverage constraints \citep{Tarpanelli2022-ac}.

While multi-sensor approaches using both SAR and MS data can potentially overcome single-sensor limitations \citep{hamidi2023fast, tarpanelli2022effectiveness}, operational reality presents significant challenges. The simultaneous availability of both sensors during critical flood periods remains severely limited, necessitating robust methodologies that function effectively with incomplete sensor combinations.
Current flood mapping methods typically require specific sensor combinations for optimal performance, yet real-world scenarios frequently present incomplete data availability. This creates a critical operational gap where emergency managers must generate actionable intelligence from whatever sensor data becomes available first, highlighting the urgent need for sensor-flexible methodologies.

Previous studies have explored deep learning techniques for multi-sensor flood mapping \citep{zhu2017deep,katiyar2021near, konapala2021exploring}. Drakonakis et al. (2023) developed OmbriaNet, demonstrating enhanced robustness when simulating realistic scenarios with missing sensor inputs \citep{drakonakis2022ombrianet}. However, current approaches face critical limitations: they require separate models for different sensor combinations with substantial computational overhead, and rely on training from scratch without leveraging pre-trained representations, necessitating large labeled datasets that are expensive to create \citep{sun2019can}.
Foundation models for Earth observation have shown considerable potential by accelerating fine-tuning for downstream tasks with limited labeled data. Prithvi \citep{jakubik2023foundation}, based on a masked autoencoder architecture and pre-trained on the Harmonized Landsat and Sentinel-2 (HLS) dataset, has shown success in flood mapping but relies on single-modality MS data, making it vulnerable to cloud cover during disasters. In contrast, Presto \citep{tseng2023lightweight}, a parameter-efficient, pixel-based model pre-trained on both S1 and S2 data, processes individual pixel-timeseries and enables channel-level masking, allowing inference when certain sensor channels are unavailable. This capability enables dynamic adaptation between SAR-only and MS-only scenarios without requiring separate models. However, Presto's application to flood inundation mapping remains unexplored.

This study introduces a novel flood inundation mapping methodology using Presto, designed to detect floods despite missing sensor data while maintaining high performance across various operational scenarios. Our research addresses critical operational flood mapping gaps by enabling diverse sensor configurations through a single model architecture, providing essential capabilities for real-world disaster scenarios requiring rapid response with available sensor data.

\section*{2 Methodology}

\begin{figure}
    \centering
    \includegraphics[width=\linewidth]{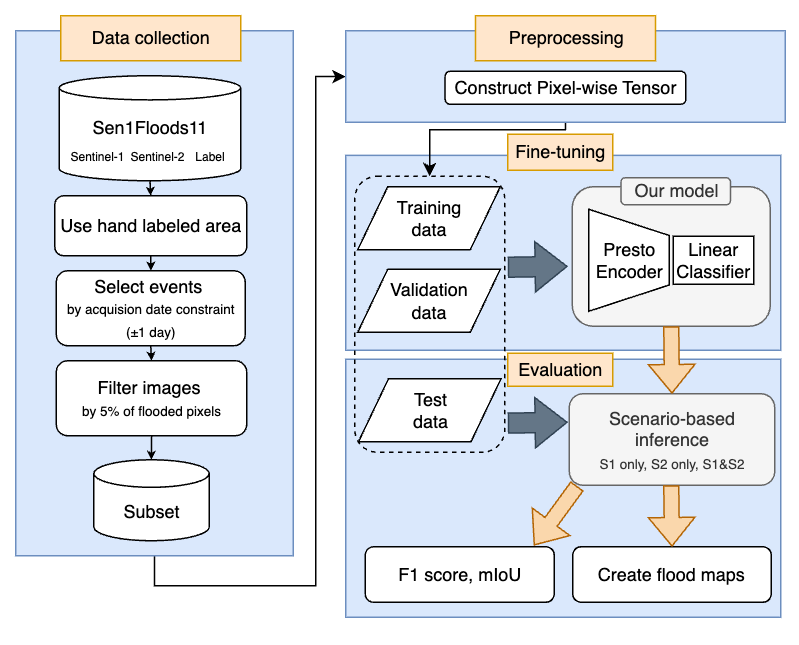}
    \caption{Overall workflow of the proposed method.}
    \label{fig:flowchart}
\end{figure}

Figure \ref{fig:flowchart} illustrates the overall workflow of our proposed method. 
Our approach accepts S1, S2, or both S1 and S2 observation data captured during flooding as input to estimate flood extents at the pixel level.

\subsection*{2.1 Model Architecture}

Our proposed model architecture leverages the pre-trained Presto encoder \citep{tseng2023lightweight} for processing a rich set of multi-modal data, complemented by a lightweight classification head for efficient pixel-wise prediction. 
The encoder features a dynamic-in-time component with 15 continuous channels and one categorical channel per monthly time step. The continuous channels include Sentinel-1 data (VV and VH backscatter), Sentinel-2 multispectral bands (B2, B3, B4, B5, B6, B7, B8, B8A, B11, and B12), the Normalized Difference Vegetation Index (NDVI), and ERA5 climate variables (temperature and precipitation). The categorical channel represents land cover class from Dynamic World \citep{brown2022dynamic}. The static component uses five channels: topography (elevation and slope) and geographic location (3D Cartesian coordinates).

The encoder processes pixel-wise time series data by transforming channel-group values from each pixel into high-dimensional feature representations. Input tokens are enhanced with positional, temporal (month), and channel-group encodings before moving through multiple transformer layers. For a sequence of input pixels, the encoder produces a final output dimension of 128.

A dedicated classification head is appended to the encoder to perform pixel-wise flood classification. Unlike approaches that aggregate features (e.g., via mean-pooling) to produce a single prediction, our head applies a linear transformation to each pixel's feature vector in the encoder's output sequence.
Specifically, the head consists of a single fully connected layer that maps the encoder's 128-dimensional feature vector for each pixel to a single logit representing the probability of flooding. 
The final binary flood map was generated using a threshold of 0.5. 

Our model accepts inputs from two channel groups of S1 and S2. A single time step of data corresponding to the flood occurrence, composed of 13 channels from S1, S2, and NDVI, which are similar to the inputs of Presto, is used for flood mapping. We achieved this through the masking strategy described below.

\subsection*{2.2 Data Collection and Preprocessing}

For our model training and evaluation, we curated a specific subset from the Sen1Floods11 dataset \citep{bonafilia2020sen1floods11}, which is designed for flood extent detection. We selected only flood events where the S1 and S2 acquisition dates differed by no more than one day, ensuring temporal consistency between the multi-sensor data. Additionally, we used hand-labeled data; it is important to note that, according to the dataset's methodology, these ground truth labels were created by analysts who primarily corrected a reference water classification derived from S2 optical imagery \citep{bonafilia2020sen1floods11}. We then filtered out image chips with flood pixel ratios below 5\% to mitigate class imbalance and ensure effective learning. Our final dataset comprises six flood events distributed across training, validation, and test sets, as detailed in Table \ref{tab:dataset_overview}.

\begin{table}[h]
    \centering
    \caption{Overview of the subset data used for fine-tuning and evaluation. Each patch is 512x512 pixels.}
    \label{tab:dataset_overview}
    \begin{tabular}{|l|c|c|c|c|c|}
    \hline
    \multirow{2}{*}{\textbf{Location}} & \multirow{2}{*}{\textbf{S1 Date}} & \multirow{2}{*}{\textbf{S2 Date}} & \multicolumn{3}{c|}{\textbf{Number of Patches}} \\ \cline{4-6} 
    & & & \textbf{Train} & \textbf{Val} & \textbf{Test} \\ \hline
    Ghana & 2018-09-18 & 2018-09-19 & 3 & 2 & 0 \\ \hline
    India & 2016-08-12 & 2016-08-12 & 18 & 7 & 0 \\ \hline
    Pakistan & 2017-06-28 & 2017-06-28 & 3 & 2 & 0 \\ \hline
    Paraguay & 2018-10-31 & 2018-10-31 & 14 & 7 & 0 \\ \hline
    USA & 2019-05-22 & 2019-05-22 & 12 & 2 & 0 \\ \hline \hline
    Bolivia & 2018-02-15 & 2018-02-15 & 0 & 0 & 15 \\ \hline
    \hline
    \textbf{Total} & \multicolumn{2}{c|}{-} & \textbf{50} & \textbf{20} & \textbf{15} \\ \hline
    \end{tabular}
\end{table}

\subsection*{2.3 Training Process}

We fine-tuned the pretrained Presto model using a curated subset of the Sen1Floods11 dataset. Our training utilized 50 chips (approximately 13 million pixels) and 20 validation chips (5.2 million pixels), with each pixel treated as an independent sample. The fine-tuning process was end-to-end, updating weights in both the Presto encoder and our randomly initialized classification head. While we made the core transformer layers and channel-specific embedding layers trainable, we kept the positional and month embeddings frozen, following the original model's fine-tuning design.

The Presto encoder accepts inputs from S1 alone, S2 alone, or both S1 and S2 observation data. This flexibility comes from channel-group masking during pre-training, which systematically removes entire sensor modalities to ensure robust performance across varying data availability scenarios.
Our model was trained exclusively on complete 13-channel sensor-fused data (MS+SAR), without implementing the structured masking strategies used in Presto's pre-training phase. During pre-training, Presto employs various masking techniques including channel-group masking that removes entire sensor modalities. For MS and SAR data, this approach systematically removes either all S1 or all S2 related bands, forcing the model to reconstruct missing sensor information from available data.

In the fine-tuning phase, we deliberately used complete sensor-fused inputs during training to maximize information content for the supervised task. As a result, only parameters corresponding to the utilized channel groups and the classification head were updated through backpropagation. Embedding layers for unused channel groups (such as ERA5 and Dynamic World) received no gradients and retained their original pre-trained weights. We applied the channel masking technique, which enables sensor-independent inference by removing unused input tokens, only during evaluation to test the model's robustness to incomplete sensor data.

To address the class imbalance between flooded and non-flooded pixels, we employed the focal loss function \citep{lin2017focal}. The model optimization was performed using the $AdamW$ optimizer with a learning rate of $1 \times 10^{-4}$ and a batch size of 4096. The entire training process was conducted on a single NVIDIA GeForce RTX 4090 GPU and completed in approximately 23.5 hours.

\section*{3. Experimental Setup}
\subsection*{3.1 Test Scenarios}

We used a test set consisting of 15 chips from a flood event in Bolivia. This region was geographically separate from the training and validation datasets to assess the model's generalization capabilities.

We designed our experiments to evaluate the proposed model's effectiveness in both ideal sensor-fusion settings and realistic operational conditions where data availability is often limited. During flood events, rapid response is critical, but observations from satellites are frequently unavailable. Even when MS images are captured, cloud cover often renders the terrestrial surface invisible. To systematically evaluate our model's performance under these constraints, we tested three scenarios:
\begin{itemize}
    \item \textbf{SAR-only}: Using the 2 channels from S1 to simulate conditions where S2 data is unavailable due to heavy cloud cover or missing observations.
    \item \textbf{MS-only}: Using the 11 channels from S2 and NDVI to simulate clear-weather conditions where SAR data are unavailable.
    \item \textbf{MS+SAR}: Using the all 13 channels from both S1 and S2 with NDVI data to assess the performance of sensor fusion.
\end{itemize}

\subsection*{3.2 Benchmark Model}

To benchmark our model, we compared it against Prithvi-100M \citep{jakubik2023foundation}. We selected Prithvi as a relevant benchmark because it is a prominent, publicly available foundation model also fine-tuned on the Sen1Floods11 dataset for flood mapping. This enables a direct comparison between two distinct approaches: our parameter-efficient pixel-based model ($\sim$0.4M parameters) versus a large-scale ($\sim$100M parameters) patch-based one. It should be noted that Prithvi only accepts HLS data (6 channels), which prevents a direct comparison with our model. The advantage of the HLS data is its frequent observation capability every two to three days, increasing the chances of capturing flood events, though cloud cover issues remain challenging. Despite these differences, comparing our model with Prithvi provides valuable insights given the substantial contrast in model size.

\subsection*{3.3  Evaluation Metrics}

To evaluate the performance in the flood detection task, the mean intersection over union (mIoU), recall, precision, and F1 score were reported using a prediction probability threshold of 0.5. The metrics were calculated using the number of True Positives ($n_{tp}$), False Positives ($n_{fp}$), True Negatives ($n_{tn}$), and False Negatives ($n_{fn}$) in a confusion matrix. When calculating the confusion matrix for the flood class, the no-data class found in the test dataset was ignored.

The metrics are calculated as follows:
\begin{itemize}
    \item \textbf{Mean Intersection over Union (mIoU)}: The mean IoU is the average IoU across all classes and provides a comprehensive measure of segmentation performance. It is computed as:
    \begin{equation}
        mIoU = \frac{1}{2} \left( \frac{n_{tp}}{n_{tp} + n_{fp} + n_{fn}} + \frac{n_{tn}}{n_{tn} + n_{fp} + n_{fn}} \right)
    \end{equation}
 
    where $N$ is the number of classes.
    \item \textbf{Recall}: Recall measures the proportion of actual positives that are correctly identified by the model:
    \begin{equation}
        Recall = \frac{n_{tp}}{n_{tp} + n_{fn}}
    \end{equation}
    \item \textbf{Precision}: Precision measures the proportion of predicted positives that are correctly identified by the model:
    \begin{equation}
        Precision = \frac{n_{tp}}{n_{tp} + n_{fp}}
    \end{equation}
    \item \textbf{F1 Score}: The F1 score is the harmonic mean of precision and recall, providing a balance between the two metrics:
    \begin{equation}
        F1 = \frac{2 \times n_{tp}}{2 \times n_{tp} + n_{fp} + n_{fn}}
    \end{equation}
\end{itemize}

\section*{4 Results}

\subsection*{4.1 Quantitative Results}
We evaluated our approach across three sensor availability scenarios using the test dataset (Table \ref{tab:results}). In the optimal sensor fusion scenario (MS+SAR), our proposed model achieved an F1 score of 0.896 and mIoU of 0.886, outperforming all other evaluated configurations. The model demonstrated superior performance across all metrics compared to the Prithvi baseline.

The MS-only scenario demonstrated our model's robustness, with minimal performance degradation and consistently higher F1 and mIoU scores than the Prithvi baseline. This confirms the model's effective use of MS features and its ability to maintain high accuracy even without SAR data. This is an important capability when operational constraints limit data availability.

The SAR-only scenario presented the greatest challenge for flood detection in terms of quantative evaluation. While our model maintained functional capabilities with an F1 score of 0.718, it showed lower performance compared to both multi-sensor scenarios and the Prithvi baseline. This performance gap highlights the inherent difficulties of SAR-based flood mapping, including issues with speckle and the complexity of interpreting backscatter signals without complementary optical information.

\begin{table}[h]
    \centering
    \caption{Model performance with Prithvi baseline on flood mapping task under SAR-only, multispectral (MS)-only, and co-observation (MS+SAR) scenarios.}
    \begin{tabular}{|l|r|l|l|l|l|}
    \hline
        \textbf{Model} & \textbf{No. Bands} & \textbf{mIoU} & \textbf{Precision} & \textbf{Recall} & \textbf{F1} \\ \hline
        \textbf{Prithvi-100M} & 6 & 0.867 & 0.865 & 0.887 & 0.876 \\ \hline
        \textbf{Ours (SAR)} & 2 & 0.736 & 0.826 & 0.636 & 0.718 \\ \hline
        \textbf{Ours (MS)} & 10 & 0.884 & \textbf{0.881} & 0.906 & 0.893 \\ \hline
        \textbf{Ours (MS+SAR)} & 12 & \textbf{0.886} & 0.880 & \textbf{0.913} & \textbf{0.896} \\ \hline
    \end{tabular}
    \label{tab:results}
\end{table}

\subsection*{4.2 Qualitative Analysis}
\begin{figure*}[htbp]
    \centering
    \includegraphics[width=\linewidth]{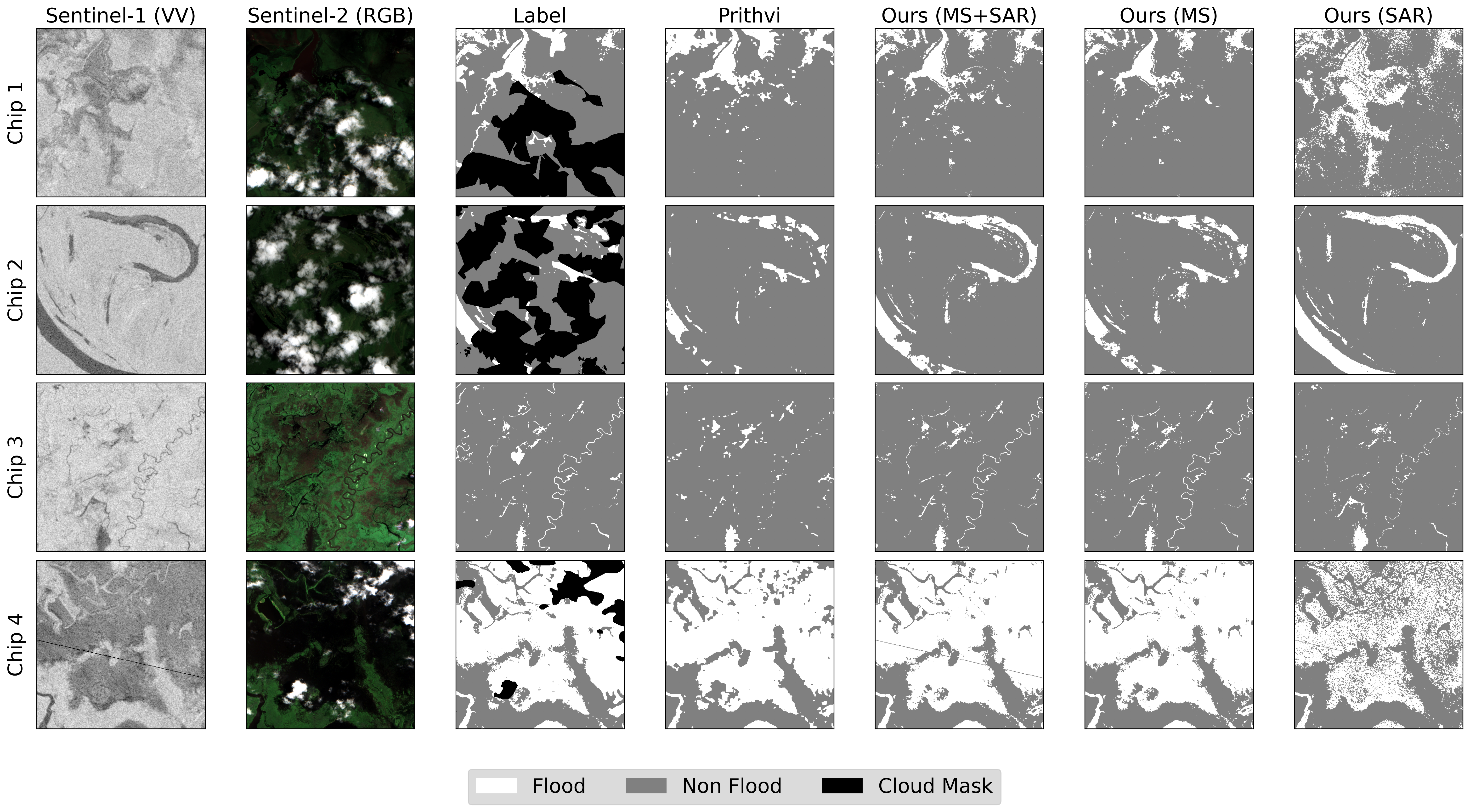}
    \caption{Examples of flood mapping on test data. Each row shows the chips of Bolivia test data. The columns represent the Sentinel-1 (SAR) data, Sentinel-2 (MS) data, ground truth, and model prediction, respectively.}
    \label{fig:results}
\end{figure*}

The selected test chips highlight a key advantage of our pixel-based approach over the patch-based Prithvi baseline (Figure \ref{fig:results}). In Chip 1 and Chip 3, our MS+SAR and MS-only models more accurately delineated the fine water extends and successfully identified narrow river channels that the Prithvi model missed.

The results also revealed trade-offs between different sensor-input scenarios. While scenarios including MS data (MS+SAR and MS-only) generally yielded more precise segmentation results, the SAR-only model demonstrated crucial utility in clouded scenes. In Chip 2, where clouds obscured the optical view, the SAR-only model identified potential water bodies not detected by MS-based approaches. This advantage was understated in the quantitative assessments because reference labels are produced based on S2 scenes, where actual flood extents are hidden by cloud cover.

Conversely, the SAR-only results exhibit certain limitations. Speckle affects the detection of large, open water bodies (Chips 1, 4), and there is a tendency to misclassify other low-backscatter surfaces as water. Additionally, a processing artifact visible in the S1 observation of Chip 4 impacts the predictions in SAR-inclusive scenarios.

\section*{5 Discussion}
A critical operational challenge in flood mapping is detecting flood extent as quickly as possible using whatever sensor data becomes available first. To address this challenge, we developed a sensor-flexible flood detection model that can process MS, SAR, or both MS and SAR inputs. 
We leveraged the Presto encoder, which accepts flexible combinations of inputs, and fine-tuned it with a linear head to detect floods at the pixel level. Our approach achieved robust and sensor-independent flood mapping capabilities that outperform a large-scale, single-modality baseline. The ability to maintain performance even with single modality data (SAR-only or MS-only scenarios) provides a substantial advantage for real-world disaster response.

However, some limitations remain. One of the challenge is the quantitative performance degradation observed in the SAR-only scenario. We hypothesize that this discrepancy may not solely reflect the model's capability but could also be attributed to a potential bias in the Sen1Floods11 ground truth labels, derived from S2 imagery. This reliance on MS data could introduce challenges for a model trained or evaluated on SAR data. For instance, water bodies obscured by clouds in the S2 image would be unlabeled in the ground truth; a SAR-based model that identifies them would thus be unfairly penalized during evaluation. Therefore, the lower evaluations in the SAR-only scenario might reflect this inherent bias in the label generation process.

Furthermore, a qualitative review of the prediction maps (Figure \ref{fig:results}) highlighted a characteristic of the pixel-wise approach. In the SAR-only scenario, the resulting flood segments exhibit salt-and-pepper noise internally. This is an expected artifact of the model classifying each pixel independently. However, it is noteworthy that despite this internal noise, the model successfully delineates the overall boundaries of the inundated areas regardless of cloud cover. This suggests that the model's core detection capability is advantageous, and the final map's quality could be enhanced with a simple post-processing step, such as a spatial smoothing filter. For future work, a promising direction to address the modality mismatch identified in the discussion would be to fine-tune the model using SAR-based ground truth labels for SAR observations. This approach holds the potential to resolve the discrepancy between the input data and the labels, thereby improving performance in the SAR-only scenario and enhancing the model's operational utility in all-weather conditions.

\section*{6 Conclusions}

This study addressed the critical need for a robust and sensor-independent flood inundation mapping method, particularly for challenging real-world disaster scenarios with varying sensor availability. We introduced a novel approach by fine-tuning Presto, a pixel-based, multi-modal pre-trained transformer model, for flood mapping tasks.

Our experiments demonstrated that this parameter-efficient model not only effectively fuses Sentinel-1 (SAR) and Sentinel-2 (MS) data but also quantitatively outperforms a popular large-scale foundational model, Prithvi-100M, in both the co-observation (SAR+MS) and MS-only scenarios. The model successfully delineated flood extents even when relying on a single sensor type (SAR-only or MS-only). Our approach has proven beneficial for monitoring real-world flood events, maximizing the use of available sensors during emergency scenarios.

\section*{Funding}
This study is a part of the Disaster
Prevention Research Institute’s Implementation Science Research for Regional
Communities (Specific) project at  Kyoto University (Project No. 2024RS-01) and NEDO (23200859-0).


\bibliographystyle{plainnat}
\bibliography{biblio}





\end{document}